\documentclass[runningheads]{llncs}

\usepackage{eccv}

\newcommand{\charimage}[1]{\raisebox{-0.25\height}{\includegraphics[height=\baselineskip]{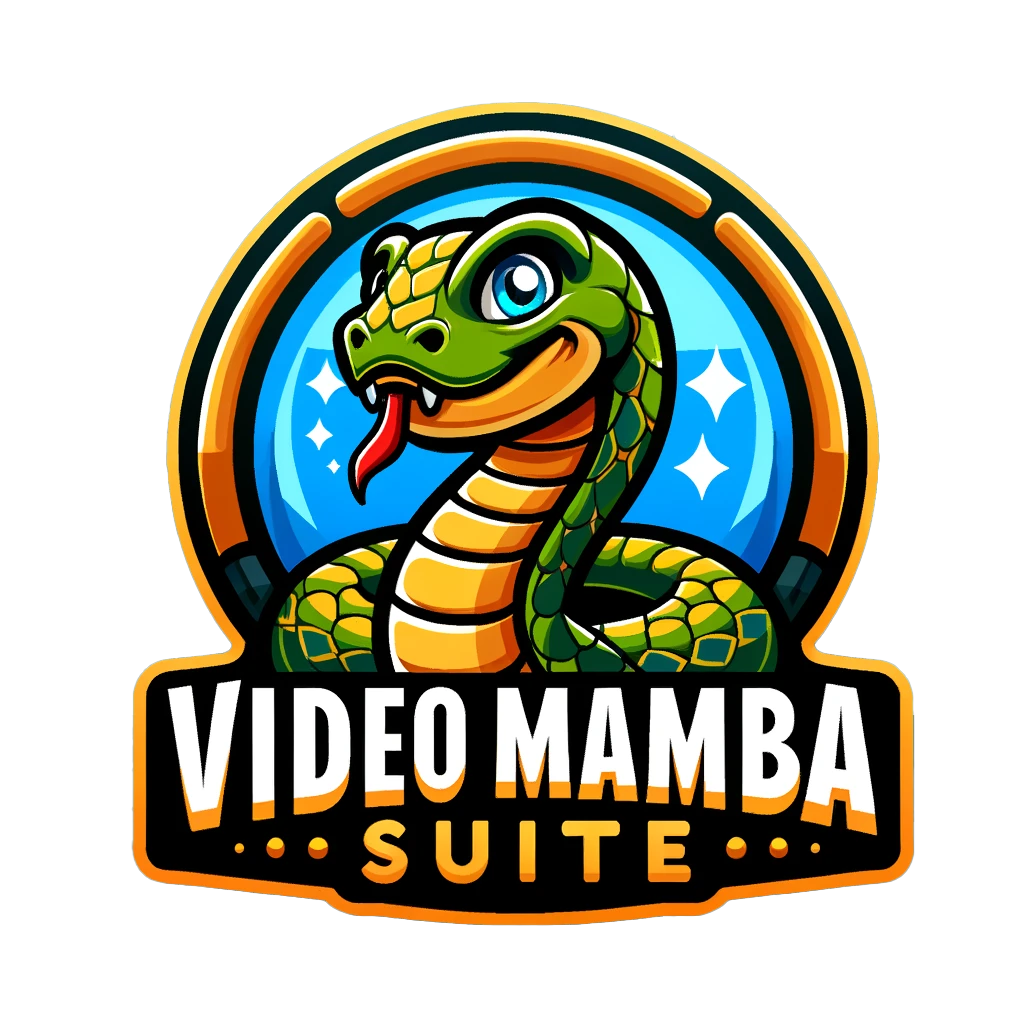}}}

\usepackage{eccvabbrv}

\usepackage{soul, color, xcolor}
\usepackage{colortbl, booktabs}
\usepackage{multirow}
\usepackage{graphicx}
\usepackage[accsupp]{axessibility}  

\usepackage[pagebackref,breaklinks,colorlinks,citecolor=ForestGreen]{hyperref}

\usepackage{orcidlink}

\begin{document}

\title{Video Mamba Suite: State Space Model as a Versatile Alternative for Video Understanding} 

\titlerunning{Video Mamba Suite}

\author{Guo Chen\inst{1}$^{\dag}$ \and
Yifei Huang\inst{2}$^\dag$ \and
Jilan Xu\inst{3}$^\dag$ \and
Baoqi Pei\inst{4,2}$^\dag$ \and
Zhe Chen\inst{1} \and
\\Zhiqi Li\inst{1} \and
Jiahao Wang\inst{1} \and
Kunchang Li\inst{2} \and
Tong Lu\inst{1}$^*$ \and
Limin Wang\inst{1,2}$^*$}
\authorrunning{G.~Chen et al.}

\institute{$^1$Nanjing University $\quad$ $^2$OpenGVLab, Shanghai AI Laboratory \\  $^3$Fudan University $\qquad$ $^4$Zhejiang University}

\maketitle

\let\thefootnote\relax\footnotetext{ $^\dag$ Equal key contributions by Guo Chen, Yifei Huang, Jilan Xu, and Baoqi Pei. \\ $^*$ Tong Lu and Limin Wang are joint corresponding authors.}


\vspace{-1.5em}
\begin{abstract}
Understanding videos is one of the fundamental directions in computer vision research, with extensive efforts dedicated to exploring various architectures such as RNN, 3D CNN, and Transformers. 
The newly proposed architecture of state space model, \textit{e.g.}, Mamba, shows promising traits to extend its success in long sequence modeling to video modeling. 
To assess whether Mamba can be a viable alternative to Transformers in the video understanding domain, in this work, we conduct a comprehensive set of studies, probing different roles Mamba can play in modeling videos, while investigating diverse tasks where Mamba could exhibit superiority. 
We categorize Mamba into four roles for modeling videos, deriving a \texttt{Video Mamba Suite}\charimage{example-image} composed of 14 models/modules, and evaluating them on 12 video understanding tasks. Our extensive experiments reveal the strong potential of Mamba on both video-only and video-language tasks while showing promising efficiency-performance trade-offs.
We hope this work could provide valuable data points and insights for future research on video understanding. Code is public: \url{https://github.com/OpenGVLab/video-mamba-suite}.
\end{abstract}

\begin{figure}[t]
    \centering
    \includegraphics[width=\textwidth]{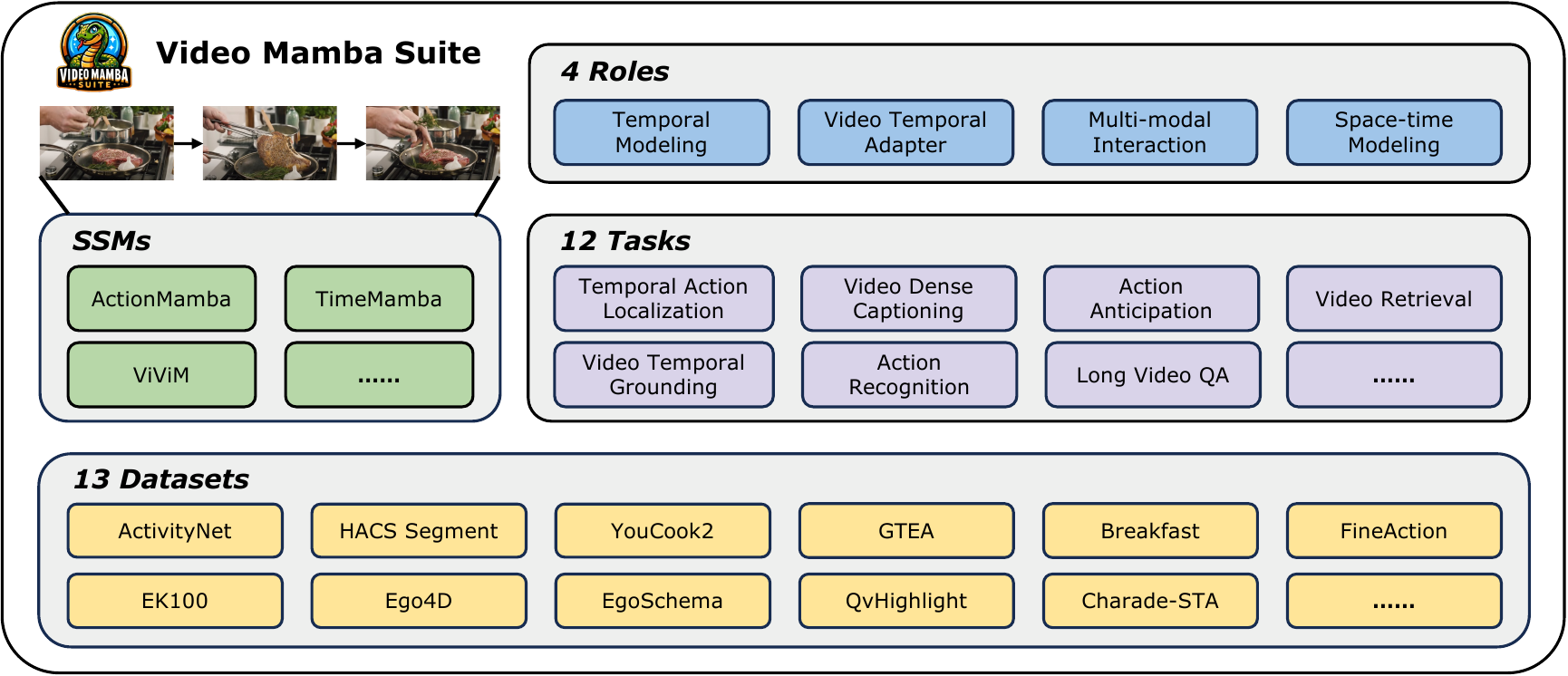}
    \vspace{-1.5em}
    \caption{We investigate SSMs exemplified by Mamba on video understanding. Our \texttt{Video Mamba Suite} comprises 14 SSM models/modules for 12 video understanding tasks. We explore 4 roles of SSM in video modeling and conduct extensive experiments on 13 major datasets.}
    \label{fig:suite}
    \vspace{-2em}
\end{figure}

\section{Introduction}
\label{sec:intro}

Video understanding is a foundation problem in computer vision research, which requires capturing spatial-temporal dynamics from videos to localize activities or infer their evolution. 
Current explorations on architectures for video understanding can be divided into three categories. The first line of work employs frame-based feature encoding followed by temporal dependency modeling via recurrent networks such as GRU and LSTM. Since this type of divided space-time modeling cannot capture joint spatiotemporal information, another stripe of work uses 3D kernels in convolutional neural networks to simultaneously consider spatial and temporal correlations~\cite{slowfast,r3d}.

Following the triumph of language~\cite{llama,gpt2,t5,transformer} and image~\cite{vit,clip,deit,beit,mae} transformers, video transformers~\cite{videomae,vivit,wang2023videomaev2} have also made significant strides in video understanding, exhibiting stronger capabilities compared to RNNs and 3D-CNNs. Video transformers encapsulate videos within a sequence of tokens, and the attention mechanism can enable global context interaction and data-dependent dynamic computation. Consequently, the model is adept at handling temporal~\cite{actionformer,asformer,wang2023mat,chen2023videollm} or spatial-temporal~\cite{videomae,timesformer} information within videos in a unified manner. 
Due to the limited computational efficiency of video transformers in long videos, several variants~\cite{video-swin,timesformer,vivit,mvit}  have emerged, balancing the speed-performance trade-off.

Recently, State Space Models (SSMs) have demonstrated their advantages in Natural Language Processing (NLP). Modern SSMs \cite{gu2021s4} exhibit strong representation ability in NLP especially in long sequence modeling, while maintaining linear-time complexity. This is because their selection mechanism can eliminate the need to store the complete context. 
Notably, Mamba \cite{gu2023mamba}, incorporates time-varying parameters into the SSM and proposes a hardware-aware algorithm to enable highly efficient training and inference. The impressive scaling performance of Mamba indicates that it is a promising alternative to transformers. Meanwhile, Mamba's strong performance and efficiency make it extremely suitable for video understanding tasks. However, despite some initial attempts to explore how Mamba can be applied in image modeling~\cite{zhu2024vim, liu2024vmamba}, its effectiveness in video understanding remains unclear. 
The absence of a comprehensive study on Mamba's potential for video understanding hinders further exploration of its capabilities in the diverse range of video-related tasks.

In this paper, we do not propose a novel method. Instead, we conduct an extensive investigation into the potential of SSM exemplified by Mamba in the context of video understanding. Our objective is to assess whether Mamba can be a viable alternative to transformers in this domain. To achieve this, we explore different roles Mamba can potentially play in understanding videos, as well as investigate diverse tasks where Mamba may exhibit superiority. We categorize Mamba into four roles for modeling videos: 1) temporal model, 2) temporal module, 3) multi-modal interaction network, and 4) spatial-temporal model. For each role, we investigate its video modeling capabilities on different video understanding tasks. To ensure a fair comparison with transformers, we carefully select counterpart models based on either standard or improved transformer architectures. 
Our exploration derives a \texttt{Video Mamba Suite}\charimage{example-image}, comprising 14 models/modules for 12 video understanding tasks. We hope our \texttt{Video Mamba Suite} can serve as a valuable resource for future explorations of SSM-based models in the field of video understanding.

\section{Related Work}

\subsection{Video Modeling}
Video modeling serves as the cornerstone for achieving deep video understanding and has undergone significant development over time. Initially, TSN~\cite{tsn} employed uniform video sampling and utilized 2D networks \cite{bninception, resnet} to establish multi-frame consensus for generating video representations. After that, substantial advances in video convolutional networks have been made. Approaches like~\cite{r2+1d,tdn,tsm} focused on integrating temporal modules into 2D networks to facilitate spatial-temporal modeling. Conversely, methods such as~\cite{slowfast,r3d,i3d,p3d,s3d,csn} either augment the kernels of 2D convolutional networks~\cite{resnet,resnext,bninception} or train 3D convolutional networks from scratch to extract spatial-temporal features. 

Nevertheless, convolutional networks are constrained by their reliance on static local operators, resulting in limited representation capacity. Inspired by the effectiveness of language transformers~\cite{gpt2,gpt3,bert,t5,llama} and image transformers~\cite{vit,deit,wang2021pvtv1,swin,mae,mvit,chen2023internvl}, various structures of video transformers have been investigated by researchers \cite{timesformer,vivit,uniformer,video-swin}. Notably, the video transformer incorporating spatial-temporal joint attention demonstrates superior capabilities among these structures. Consequently, subsequent research efforts \cite{videomae,videomae_facebook,wang2023videomaev2,unmasked-teacher,wang2022internvideo} delve into different pre-training methodologies based on the structure. However, the quadratic computational complexity associated with joint attention poses a significant obstacle, impeding the scalability of video transformers for handling longer contexts, which is similar to what issue LLMs \cite{touvron2023llama,touvron2023llama2} face. 
Several studies have developed transformer-based variants \cite{timesformer,vivit,video-swin} tailored for modeling long-form videos frequently consisting of more than 128 frames. 

Alternatively, another approach is to design model architectures with linear complexity. For instance, RetNet~\cite{sun2023retnet} and RWKV~\cite{peng2023rwkv} utilized exponential decay techniques to capture global information. State space models~\cite{gu2021s4, gu2021combining, gu2022parameterization} also offer linear complexity, with Mamba~\cite{gu2023mamba} employing efficient implementations to facilitate data-dependent inference. In the vision field, some works~\cite{ali2021xcit,zhu2024vim,duan2024vision-rwkv} have explored vision encoders with linear complexity.
XCiT~\cite{ali2021xcit} achieved global interaction with linear complexity by calculating the cross-variance between input tokens. 
Recently, \cite{zhu2024vim,liu2024vmamba} has conducted an initial exploration of a Mamba-based vision application.
In our study, we investigate the creation of a Mamba-based video model for video understanding.

\subsection{State-Space Models (SSMs)}

As a frontrunner in the era of State-Space Models (SSMs), \cite{gu2021s4} introduced a novel model called the Structured State-Space Sequence (S4), which offers an alternative to CNNs and transformers for capturing long-range dependencies. 
The S4 model exhibits a promising characteristic of linearly scaling with sequence length.
Building on this, \cite{smith2022s5} proposed an advanced layer dubbed S5, which integrates MIMO SSM with efficient parallel scanning into the S4 architecture. This development seeks to overcome the constraints of SSMs, enhancing their efficacy.
Furthermore, \cite{fu2022h3} contributed a novel SSM layer, H3, significantly closing the performance gap between SSMs and transformer-based attention in language modeling. 
\cite{mehta2022gss} extended the S4 model by introducing additional gating units in the Gated State Space layer to enhance its expressivity. 
More recently, \cite{gu2023mamba} introduced a data-dependent SSM layer and developed a versatile language model called Mamba. Mamba outperforms transformers in performance across various sizes of large-scale real data and demonstrates linear scaling with sequence length. 
In this study, we aim to extend the success of Mamba in the language domain to video understanding. 
Specifically, we construct a versatile framework, termed \texttt{Video Mamba Suite}\charimage{example-image}, to develop, validate, and analyze the performance of Mamba for video understanding.

There are only limited works that use SSMs for video understanding, with the primary focus on long-term video classification \cite{wang2023lsmcl,islam2022vis4mer}. \cite{islam2022vis4mer} expanded the S4 architecture and incorporated a multi-scale temporal S4 decoder into video sequence classification. \cite{wang2023lsmcl} introduced a masked contrastive learning framework to augment state-space models. 
Our study extends the application scope of SSMs to a much broader range of video understanding tasks.
We cover multiple usages of SSMs and more video-related tasks, using Mamba~\cite{gu2023mamba} as an example, moving beyond the sole focus on the classification of long videos.

\section{Preliminaries}
\label{sec:preliminary}

\subsection{State Space Model}
In this section, we introduce models based on Structured State-Space (SSM), specifically the S4 \cite{gu2021s4} and Mamba \cite{gu2023mamba} models.
These models draw inspiration from continuous systems that process sequences or functions. 
Imagine a system that takes a sequence of inputs over time, $x(t)$, and produces a sequence of outputs, $y(t)$, while transforming the input through a hidden state, $h(t)$.
 
This system uses matrices $\mathbf{A}^{\text{N}\times \text{N}}$ to define the evolution of the hidden state, and $\mathbf{B}^{\text{N}\times \text{1}}$ and $\mathbf{C}^{\text{1}\times \text{N}}$ to project the input and hidden state to the output, respectively.
This process can be summarized as $h'(t) = \mathbf{A}h(t) + \mathbf{B}x(t),y(t) = \mathbf{C}h(t)$.

S4 and Mamba are designed as discrete equivalents of these continuous systems. 
They incorporate a timescale parameter $\Delta$, to convert the continuous parameters $(\mathbf{A}$, $\mathbf{B})$ into their discrete counterparts $(\overline{\mathbf{A}},\overline{\mathbf{B}})$.
The conversion uses a technique, zero-order hold, resulting in the discrete formulas,  $\overline{\mathbf{A}} = \text{exp}(\Delta \mathbf{A}), \overline{\mathbf{B}} = (\Delta \mathbf{A})^{-1}(\text{exp}(\Delta \mathbf{A})-\mathbf{I})\cdot \Delta \mathbf{B}$.
Once the parameters are discretized, the system's behavior at each discrete time step $t$ is given by $h_t = \overline{\mathbf{A}}h_{t-1}+\overline{\mathbf{B}}x_t,y_t = \mathbf{C}h_t$, showcasing how the hidden state updates and produces output.

Finally, the output is computed through a global convolution process, involving a structured convolutional kernel, $\overline{\mathbf{K}} =(\mathbf{C}\overline{\mathbf{B}},\mathbf{C}\overline{\mathbf{A}\mathbf{B}},...,\mathbf{C}\overline{\mathbf{A}}^{\texttt{M-1}}\overline{\mathbf{B}})$.
This kernel is constructed from the transformed projection parameters and applied across the entire input sequence $\mathbf{x}$ to produce the final output sequence $\mathbf{y}$, \ie $\mathbf{y} = \mathbf{x} * \overline{\mathbf{K}}$. Here, $\texttt{M}$ is the length of the input sequence $\mathbf{x}$.
By transforming continuous dynamics into discrete steps, S4 and Mamba models allow for the processing of sequences with complex dependencies.

\begin{figure}[t]
    \centering
    \includegraphics[width=\textwidth]{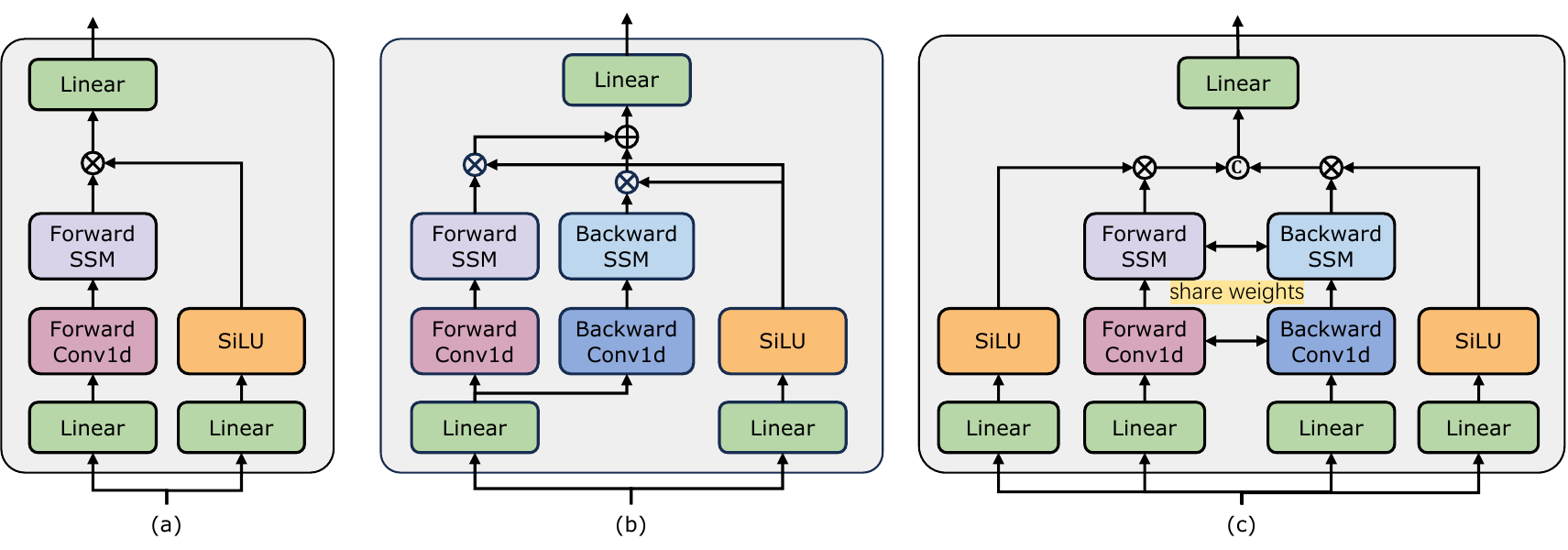}
    \vspace{-2em}
    \caption{Illustration of three SSMs blocks. (a) is the vanilla Mamba block \cite{gu2023mamba}. (b) is the ViM block \cite{zhu2024vim}. (c) is our proposed DBM block, which separates the input projector and shares the parameters of SSM in both scanning directions.}
    \label{fig:blocks}
    \vspace{-2mm}
\end{figure}

\subsection{Mamba Block}

Mamba block~\cite{gu2023mamba} combines a linear attention operator and an MLP block, inspired by the gated attention unit (GAU)~\cite{flash-gau}, as illustrated in Figure~\ref{fig:blocks} (a). 
This architecture involves expanding the model dimension \texttt{D} by a controllable expansion factor \texttt{E}. 
For each block, most of the parameters ($\texttt{3}\texttt{E}\texttt{D}^2$) are in the linear projections ($\texttt{2}\texttt{E}\texttt{D}^2$ for input projections, $\texttt{E}\texttt{D}^2$ for output projection) while the inner SSM contributes less. The number of SSM parameters (projections for $\Delta$, $\mathbf{B}$, $\mathbf{C}$, and the matrix $\mathbf{A}$) are much smaller in comparison.

\subsection{ViM Block}

The ViM block~\cite{zhu2024vim}, as shown in Figure~\ref{fig:blocks} (b), adds a backward selective scanning branch with additional parameters, to the Mamba block. Based on this, features for both scanning directions share the same linear projection and gated layers.
Given the input $\mathbf{x}$,
for each direction, the ViM block first applies 1D convolution to the input $\mathbf{x}$ and gets $\mathbf{x}'_o$. Then the block linearly projects $\mathbf{x}'_o$ to $\mathbf{B}_o, \mathbf{C}_o,\Delta_o$, respectively. $\Delta_o$ is then used to transform the $\overline{\mathbf{A}}_o$ and $\overline{\mathbf{B}}_o$, respectively. Next, SSM computes forward-scanned features $\mathbf{y}_f$ and backward-scanned features $\mathbf{y}_b$. Finally, both features are gated by the gated layer and averaged to get the output token sequence.

\subsection{DBM Block}
We further investigate a Decomposed Bidirectionally Mamba block, denoted as DBM, as depicted in Figure~\ref{fig:blocks} (c). It employs an inverse design compared to the ViM block. Given the input sequence $\mathbf{x}$, the DBM block initially employs distinct linear layers to segregate forward features $\mathbf{x}_f$ and backward features $\mathbf{x}_b$. These features are then passed through the SSM module with shared parameters for bidirectional scanning, resulting in $\mathbf{x}^{'}_{f}$ and $\mathbf{x}^{'}_{b}$. Subsequently, both features undergo gating by two distinct layers before being concatenated to generate the output token sequence. 
This block introduces directional bias and weakens specific dynamics. 
In our experiments detailed in Section~\ref{sec:vtm}, we observe that the DBM block demonstrates improved performance on small-scale datasets.

\section{Video Mamba Suite}
In this section, we present \texttt{Video Mamba Suite}\charimage{example-image}, which encompasses various Mamba-based models for a diverse set of video understanding tasks. The suite leverages Mamba in four distinct roles: temporal models, temporal modules, multi-modal interaction models, and space-time sequence models. Each role will be discussed in a separate subsection, gradually revealing Mamba's performance in video understanding and highlighting its key characteristics.
Due to space limitations, more implementation details and experimental results are provided in the supplementary material.

\subsection{Mamba for Video Temporal Modeling}
\label{sec:vtm}

\begin{table}[t]

\centering
\scriptsize
\setlength{\tabcolsep}{2.1mm}{
\begin{tabular}{lccccc} 
\toprule
Method   & Block  & mAP@0.5 & mAP@0.75 & mAP@0.95 & mAP@Avg \\
\midrule

ActionFormer~\cite{actionformer} & Window Attn~\cite{choromanski2020rethink-attn} & 62.62 & 44.61 & 12.73 & 43.34   \\
ActionMamba~\cite{zhu2024vim} & ViM~\cite{zhu2024vim} & 63.78 & 45.45 & 13.01 & 44.26   \\
\rowcolor{gray!15}
ActionMamba & DBM (ours) &  \textbf{64.02} & \textbf{45.71} & \textbf{13.34} &   \textbf{44.56} \\
\bottomrule
\end{tabular}
}
\vspace{0.5em}
\caption{
Results of temporal action localization on HACS Segment~\cite{hacs}. The metric is mean Average Precision (mAP) under multiple tIoU thresholds \{0.5, 0.75, 0.95\}.
}
\vspace{-1em}
\label{table:tal-res}
\end{table}

\begin{table}[t]
\centering
\scriptsize
\setlength{\tabcolsep}{2.3mm}{
\begin{tabular}{lcccccc} 
\toprule
Method   & Block  & Acc & Edit & F1@10 & F1@25 & F1@50 \\
\midrule
MS-TCN~\cite{ms-tcn} & Dilated Conv, Enc & 76.3 & 79.0 & 85.8 & 83.4 & 69.8   \\
\midrule
ASFormer~\cite{asformer} & Window Attn~\cite{choromanski2020rethink-attn}, Enc-Dec & \textbf{79.7} & 84.6 & 90.1 & 88.8 & 79.2   \\
ASFormer$^\dagger$~\cite{asformer} & Window Attn~\cite{choromanski2020rethink-attn}, Enc-Dec & 77.1 & 81.0 & 86.2 & 84.8 & 77.0   \\
ASFormer$^\dagger$~\cite{asformer} & Window Attn~\cite{choromanski2020rethink-attn}, Enc & 75.4 & 78.1 & 82.7 & 80.5 & 68.4   \\
ASMamba~\cite{zhu2024vim} & ViM~\cite{zhu2024vim}, Enc & 79.3 & 87.0 & 90.3 & 89.0 & 77.9   \\
\rowcolor{gray!15}
ASMamba & DBM (ours), Enc & 78.4 & \textbf{87.5} & \textbf{91.1} & \textbf{89.8} & \textbf{79.7}   \\
\bottomrule
\end{tabular}
}
\vspace{0.5em}
\caption{
Results of temporal action segmentation on GTEA~\cite{fathi2011gtea} dataset. 
The metrics are accuracy, edit distance~\cite{edit-distance}, and instance-wise F1 under multiple tIoU thresholds \{0.1, 0.25, 0.5\}. $\dagger$ denotes our reproduced results with its official code.}
\vspace{-1em}
\label{table:tas-res}
\end{table}

\subsubsection{Tasks and datasets.} We assess Mamba's performance across five video temporal tasks: temporal action localization (HACS Segment~\cite{hacs}), temporal action segmentation (GTEA~\cite{fathi2011gtea}), dense video captioning (ActivityNet~\cite{anet}, YouCook \cite{zhou2018youcook2}), video paragraph captioning (ActivityNet \cite{anet}, YouCook \cite{zhou2018youcook2}), and action anticipation (Epic-Kitchen-100~\cite{epic-Kitchens}). 

\vspace{-1em}
\subsubsection{Baseline and competitor.} We select transformer-based counterparts as the baseline of each task. Specifically, the transformer baselines are ActionFormer~\cite{actionformer}, ASFormer~\cite{asformer}, Testra~\cite{testra}, and PDVC~\cite{wang2021pdvc}.
To build a Mamba challenger, we replace transformer blocks of the baseline model with Mamba-based blocks, including vanilla Mamba~\cite{gu2023mamba}, ViM~\cite{zhu2024vim}, and our DBM. 
Note that in the context of action anticipation, which involves causal inference, we compare the performance of the baseline with the vanilla Mamba block~\cite{gu2023mamba}.

\vspace{-1em}
\subsubsection{Results and analysis.} We present the comparison results of different models for four tasks in Table~\ref{table:tal-res} to Table~\ref{table:aa-res}. Overall speaking, while some transformer-based models have incorporated attention variants to enhance performance, the tables demonstrate the superior performance of the Mamba series compared to existing methods in the transformer series.

\emph{Temporal action localization.} As depicted in Table~\ref{table:tal-res}, our introduced ActionMamba, which is based on changing Transformer blocks of ActionFormer~\cite{actionformer} to ViM blocks, achieves an average mAP of 44.26 on the HACS Segment dataset. Using our DBM blocks, ActionMamba can further improve the average mAP to 44.56. This significantly outperforms its transformer counterpart by 1.22 (44.56 \vs 43.34).

\emph{Temporal action segmentation.} Table~\ref{table:tas-res} reports the results of temporal action segmentation. Our proposed ASMamba demonstrates superior performance compared to our reproduced ASFormer~\cite{asformer}.
Since ASMamba is a pure encoder while ASFormer adopts the encoder-decoder structure, we also experiment with an encoder version of ASFormer, where a drop in performance is observed. The results suggest the great potential of Mamba compared with the Transformer.

\begin{table}[t]
\centering
\scriptsize
\setlength{\tabcolsep}{1.7mm}{
\begin{tabular}{lccccccccc} 
\toprule
  \multirow{2}{*}{Method} & \multirow{2}{*}{Block}  &  \multicolumn{4}{c}{ActivityNet} &  \multicolumn{4}{c}{YouCook2} \\
   \cmidrule(lr){3-6}\cmidrule(lr){7-10}
   & & B-4 & M & C & SODA  & B-4 & M & C & SODA \\
\midrule
PDVC~\cite{wang2021pdvc} & DeformAttn~\cite{deformable-detr} & 1.75 & 6.73 & 26.07 & \textbf{5.47} & 0.73 & 4.25 & 20.48 & 4.02  \\
PDVC~\cite{wang2021pdvc} & ViM~\cite{zhu2024vim} & 1.68 & 6.92 & 26.26 & 5.33 & 0.71 & 4.32 & 20.59 & 4.09 \\
\rowcolor{gray!15}
PDVC & DBM (ours) & \textbf{1.76} & \textbf{7.16} &\textbf{ 26.77} & 5.27 & \textbf{0.86 }& \textbf{4.44} & \textbf{22.11} & \textbf{4.32 }\\
\bottomrule
\end{tabular}
}
\vspace{0.5em}
\caption{
Results of dense video captioning on ActivityNet~\cite{anet} and YouCook2~\cite{zhou2018youcook2}. 
The metrics include BLEU-4~\cite{papineni2002bleu}, METEOR~\cite{banerjee2005meteor}, CIDEr~\cite{vedantam2015cider} and SODA\_c~\cite{fujita2020soda}.
}
\vspace{-1em}
\label{table:dvc-res}
\end{table}

\begin{table}[t]
\centering
\scriptsize
\setlength{\tabcolsep}{1.65mm}{
\begin{tabular}{lccccccccc} 
\toprule
  \multirow{2}{*}{Method} & \multirow{2}{*}{Block}  &  \multicolumn{4}{c}{ActivityNet} &  \multicolumn{4}{c}{YouCook2} \\
   \cmidrule(lr){3-6}\cmidrule(lr){7-10}
   & & B-4 & M & R & C  & B-4 & M & R & C \\
\midrule
PDVC~\cite{wang2021pdvc} & DeformAttn~\cite{deformable-detr} & 9.07 & 12.52 & 29.02 & 13.12 & 6.44 & 12.94 & 28.74 & 12.48  \\
PDVC~\cite{wang2021pdvc} & ViM~\cite{zhu2024vim} & \textbf{9.33 }& 13.52 & 29.83 & 14.25 & 6.50 & 12.96 & 28.59 & \textbf{13.08} \\
\rowcolor{gray!15}
PDVC & DBM (ours) & 9.05 & \textbf{14.05} & \textbf{29.86 }& \textbf{14.43} & \textbf{7.24} & \textbf{13.47} & \textbf{29.09} & 13.03 \\
\bottomrule
\end{tabular}
}
\vspace{0.5em}
\caption{
Results of video paragraph captioning on ActivityNet~\cite{anet} and YouCook2~\cite{zhou2018youcook2}. 
The metrics include BLEU-4~\cite{papineni2002bleu}, METEOR~\cite{banerjee2005meteor}, ROUGE-L~\cite{rouge}, CIDEr~\cite{vedantam2015cider}.
}
\vspace{-1em}
\label{table:vpc-res}
\end{table}

\emph{Dense video captioning.}
Table~\ref{table:dvc-res} shows the results on dense video captioning. 
The baseline PDVC model~\cite{wang2021pdvc} adopts a Deformable Transformer to encode visual information. In comparison, our bidirectional Mamba with DBM block shows stronger performances on both temporal event localization and caption generation.

\emph{Video paragraph captioning.}
Table~\ref{table:vpc-res} demonstrates the result comparison on video paragraph captioning. We also adopt the PDVC as the baseline model~\cite{wang2021pdvc}, by training the model with captioning loss only. Different from the dense video captioning task where both captioning and localization are critical, video paragraph captioning merely focuses on extracting fine-grained visual information for generating captions.
Results in Table~\ref{table:vpc-res} reveal our bidirectional Mamba brings stronger feature representation for captioning, compared with the temporal deformable encoder.

\emph{Action anticipation.}
We further assess the Mamba's ability in causal modeling via the action anticipation task. By considering 5-second temporal features as input, we compare Testra~\cite{testra} with causal self-attention blocks and causal Mamba~\cite{gu2023mamba} blocks. As presented in Table~\ref{table:aa-res}, the results demonstrate the superior causal reasoning capability of Mamba, which aligns with the conclusion in text generation of Table~\ref{table:vpc-res}.

\begin{table}[t]
\centering
\scriptsize
\setlength{\tabcolsep}{1.38mm}{
\begin{tabular}{lcccccccccc} 
\toprule
 \multirow{2}{*}{Method}  &  \multirow{2}{*}{Block} &  \multicolumn{3}{c}{Overall} &  \multicolumn{3}{c}{Unseen} &  \multicolumn{3}{c}{Tail} \\
   \cmidrule(lr){3-5} \cmidrule(lr){6-8} \cmidrule(lr){9-11}
   &   & Ver & Nou & Act & Ver & Nou & Act & Ver & Nou & Act \\
\midrule
Testra$^+$~\cite{testra} & long short Attn~\cite{transformer} & 30.8 & 35.8 & 17.6 & 29.6 & 26.0 & 12.8 & 23.2 & 29.2 & 14.2\\
MAT$^+$~\cite{wang2023mat} & long short Attn~\cite{transformer} & 35.0 & 38.8 & 19.5 & 32.5 & 30.3 & 13.8 & 28.7 & 33.1 & 16.9   \\
\midrule
Testra~\cite{testra} & short Attn~\cite{transformer} & 25.1 & 30.8  & 14.1 & 24.3 & \textbf{24.5} & 10.7 & 17.4 & 23.0 & 10.9   \\
\rowcolor{gray!15}
Testra & short Mamba~\cite{gu2023mamba} & \textbf{27.9}   & \textbf{34.1} & \textbf{15.2} & \textbf{28.1} & 24.2 & \textbf{12.0} & \textbf{20.5} & \textbf{27.8} & \textbf{12.3}   \\
\bottomrule
\end{tabular}
}
\vspace{1em}
\caption{
Results of action anticipation on EK-100~\cite{epic-Kitchens}. 
Accuracy measured by class-mean recall@5(\%)
following the standard protocol. $^+$ denotes the model is trained with data augmentation. ``long'' and ``short'' denote using long- and short-term memory.
}
\vspace{-1em}
\label{table:aa-res}
\end{table}

\subsection{Mamba for Cross-Modal Interaction}

\subsubsection{Tasks and datasets.} In addition to single-modal tasks, we assess the performance of Mamba for cross-modal interaction. We first employ the video temporal grounding (VTG) task for evaluation.
The involved datasets contain QvHighlight~\cite{moment_detr_qvhighlights} and Charade-STA~\cite{gao2017charade-sta}.

\vspace{-1em}
\subsubsection{Baseline and competitor.} In this work, we use UniVTG~\cite{lin2023univtg} to create our mamba-based VTG model.
UniVTG employs a transformer as the multi-modal interaction network. Given video feature $\mathbf{V}=\{\mathbf{v}_i\}^{L_v}_{i=1}\in \mathbb{R}^{L_v\times D}$ and text feature $\mathbf{Q}=\{\mathbf{q}_j\}^{L_q}_{i=1}\in \mathbb{R}^{L_q\times D}$, we first add learnable position embeddings $\mathbf{E}^{pos}$ and modality-type embeddings $\mathbf{E}^{type}$ to each modality to retain both positional and modality information:

\begin{equation}
\begin{aligned}
    \tilde{\mathbf{V}} &= \mathbf{V}+\mathbf{E}^{pos}_{V}+\mathbf{E}^{type}_{V}, \\
    \tilde{\mathbf{Q}} &= \mathbf{Q}+\mathbf{E}^{pos}_{Q}+\mathbf{E}^{type}_{Q}.
\end{aligned}
\end{equation}

Then, the text and video tokens are concatenated to get a joint input $\mathbf{Z}=[\tilde{\mathbf{V}};\tilde{\mathbf{Q}}] \in \mathbb{R}^{L\times D}$, where $L=L_v+L_q$. Further, $\mathbf{Z}$ is fed into a multi-modal transformer encoder. Finally, the text-enhanced video features $\tilde{\mathbf{V}}^{e}$ are taken out and then fed into the prediction head. 
To create a cross-modal Mamba competitor, we stack bidirectional Mamba blocks to form a multi-modal Mamda encoder to replace the transformer baseline.

\begin{table}[t]
\centering
\scriptsize
\setlength{\tabcolsep}{0.65mm}{
\begin{tabular}{lccccccccccc} 
\toprule
    \multirow{4}{*}{Method}  & \multirow{4}{*}{Block} & \multicolumn{7}{c}{Qvhighlight}  &  \multicolumn{3}{c}{Charade-STA}   \\ 
      \cmidrule(lr){3-9} \cmidrule(lr){10-12}
      & & \multicolumn{2}{c}{R1} & \multicolumn{3}{c}{mAP}   & \multicolumn{2}{c}{HD ($\geq$ VG)}& \multicolumn{3}{c}{R1}  \\
      \cmidrule(lr){3-5} \cmidrule(lr){6-7}\cmidrule(lr){8-9}\cmidrule(lr){10-12}
 & & @0.5 & @0.7 & @0.5 & @0.75 & Avg. & mAP & HIT@1 & @0.3 & @0.5 & @0.7     \\
  \midrule
MDETR~\cite{moment_detr_qvhighlights} & - & 52.89 & 33.02 & 54.82 & 29.40 & 30.73 & 35.69 & 55.60 & 65.83 & 52.07 & 30.59  \\
UMT~\cite{liu2022umt} & - & 56.23 & 41.18 & 53.83 & 37.01 & 36.12 & 38.18 & 59.99 & - & - & -   \\

\midrule
UniVTG~\cite{lin2023univtg}  &Trans & 58.86 & 40.86 & 57.60 & 35.59 & 35.47 & 38.20 & 60.96 & \textbf{70.81} & \textbf{58.01} & 35.65   \\
UniVTG$\dagger$ &Trans & 59.87 & 44.32 & 59.09 & 40.25 & 38.48 & 39.22 & 64.71 & 68.20 & 57.26 & 34.68   \\
\midrule
\rowcolor{gray!15}
UniVTG~\cite{lin2023univtg} &ViM & 65.55 & 50.77 & 63.97 & 46.11 & 44.74 & \textbf{40.22} & 64.26 & 68.12 & 57.07 & 35.83 \\
\rowcolor{gray!15}
UniVTG~\cite{lin2023univtg} &DBM (ours) & \textbf{66.65} & \textbf{52.19} & \textbf{64.37} & \textbf{46.68} & \textbf{45.18} & 40.18 & \textbf{64.77} & 68.06 & 57.18 & \textbf{36.05}  \\
\bottomrule
\end{tabular}
}
\vspace{1em}
\caption{Results of video temporal grounding tasks on Qvhighlight~\cite{moment_detr_qvhighlights} and Charade-STA~\cite{gao2017charade-sta}. UniVTG~\cite{lin2023univtg} default adopts a 4-layer transformer for cross-modal interaction. $^\dagger$ denotes the reproduced results of a 6-layer transformer.}
\vspace{-1em}
\label{table:vtg-res}
\end{table}

\begin{figure}[t]
    \centering
    \scriptsize
    \begin{minipage}{0.45 \textwidth}
    \centering
    \setlength{\tabcolsep}{1.2mm}{
    \begin{tabular}{lcccc} 
    \toprule
    \multirow{2}{*}{Method} & \multicolumn{2}{c}{Qvhighlight} & \multicolumn{2}{c}{Charade-STA}  \\
    \cmidrule(lr){2-3}\cmidrule(lr){4-5}
    & R@0.5 &R@0.7 & R@0.5 &R@0.7 \\ 
    \midrule
    L & \textbf{66.97} &\textbf{51.23} & \textbf{56.29} & \textbf{35.99} \\ 
    R & 62.06 &44.84 & 46.88 & 25.89 \\ 
    L+R & 65.10 &50.91 & 55.65 & 34.30 \\ 
    M & 61.87 &44.06 & 46.56 & 25.65 \\ 
    \bottomrule
    \end{tabular}
    \captionof{table}{Effect of the position of text tokens in the whole token sequence.}
    \label{table:video-text-token}
    }

    \end{minipage}\hfill
    \begin{minipage}{0.52 \textwidth}
        \centering
        \includegraphics[width=0.7\textwidth]{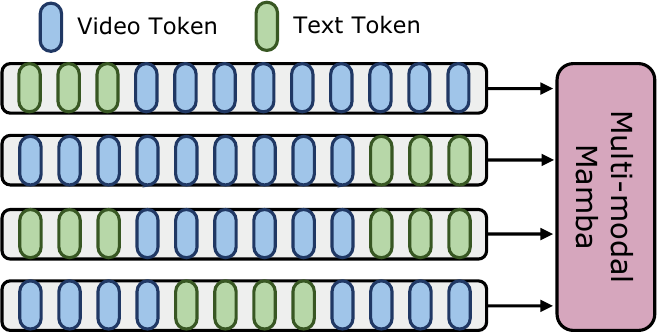}
        \caption{Illustration for different positions of video and text tokens in the input sequence.}
        \label{fig:video-text-token}
    \end{minipage}
    \vspace{-1em}
\end{figure}

\vspace{-1em}
\subsubsection{Results and analysis.}
We present the performance of multiple models on Qvhighlight~\cite{moment_detr_qvhighlights} in Table~\ref{table:vtg-res}. Mamba achieves an average mAP of 44.74, representing a significant improvement compared to the transformer (44.74 \emph{vs.} 38.48). For Charade-STA~\cite{gao2017charade-sta}, the Mamba-based method also achieves comparable performance.
This suggests that Mamba has the potential to integrate multiple modalities effectively. 
Given that Mamba~\cite{gu2023mamba} is a model based on linear scanning, while the transformer is based on global token interaction, intuitively, we believe that the position of text in the token sequence may influence the effectiveness of multi-modal aggregation. To investigate this, we include different text-visual fusion methods in Table~\ref{table:video-text-token}, while Figure~\ref{fig:video-text-token} illustrates four different arrangements of tokens. 
We observe that the best results are obtained when textual conditions are fused on the left side of the visual features. Qvhighlight~\cite{moment_detr_qvhighlights} is less affected by this fusion, whereas Charade-STA~\cite{gao2017charade-sta} shows particular sensitivity to the position of the text, which may be attributed to the characteristics of the dataset.

\begin{figure}[t]
    \centering
    \includegraphics[width=\textwidth]{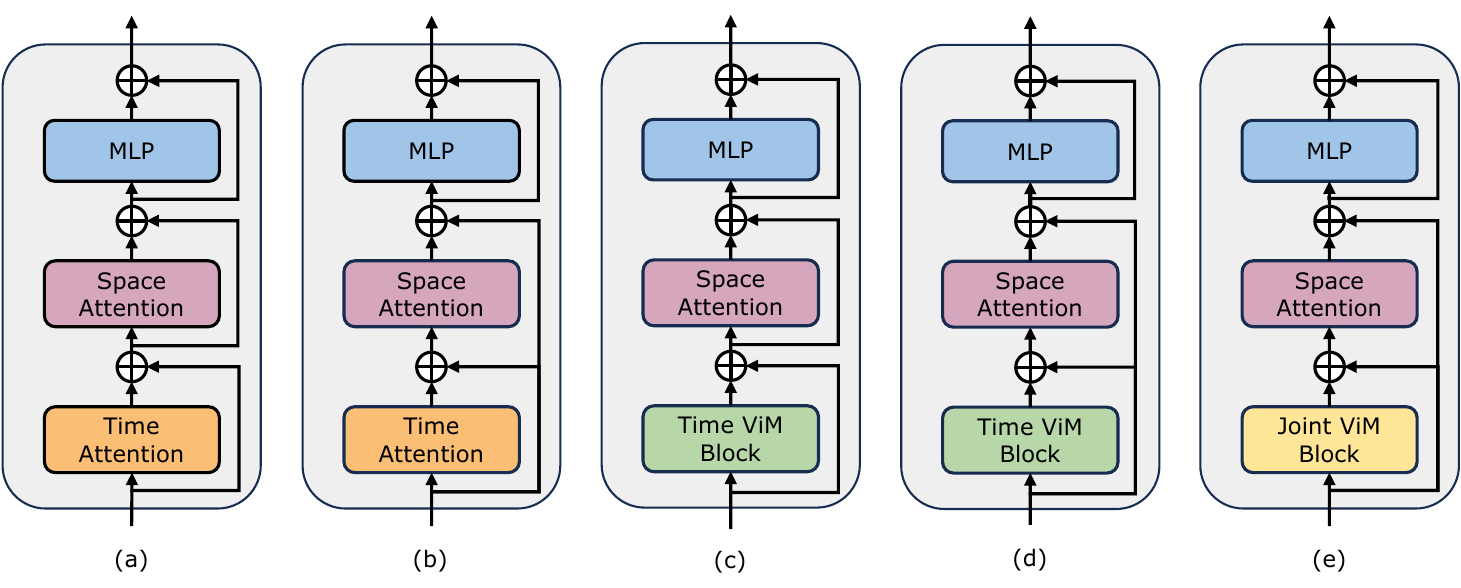}
    \vspace{-2em}
    \caption{Illustration of our explored structures. (a) and (b) shows vanilla-style~\cite{timesformer} and frozen-style~\cite{frozenintime} residual connection forms for TimeSformer~\cite{timesformer}. (c) and (d) presents our created TimeMamba which uses ViM block as a temporal module in both styles. (e) provides the replacement of the temporal ViM block with a space-time ViM block.} 
    \label{fig:adapter}
\end{figure}

\vspace{-1em}
\subsubsection{Results and analysis.}
The performance comparison on TimeMamba and TimeSformer~\cite{timesformer} is shown in  Table~\ref{table:zsmir-res}, Table~\ref{table:ft-cls-res}, Table~\ref{table:zs-qa-res} and Figure~\ref{fig:zs-qa-frame}.

\subsection{Mamba as Video Temporal Adapter}

\subsubsection{Tasks and datasets.} In addition to evaluating Mamba's performance in post-temporal modeling, we also assess its effectiveness as a video-temporal adapter. 
We pre-train a dual-tower model by performing video-text contrastive learning on the egocentric data~\cite{ego4d,lin2022egovlp}, which contains 4M video clips with fine-grained narrations. 
For evaluation, we consider zero-shot/fine-tuned multi-instance retrieval and fine-tuned action recognition on the Epic-Kitchens-100 dataset~\cite{epic-Kitchens}, and zero-shot long-form Question Answering on the EgoSchema dataset~\cite{mangalam2024egoschema}.

\vspace{-1em}
\subsubsection{Baseline and competitor.} 
TimeSformer~\cite{timesformer} adopts the divided space-time attention block to model the spatial and temporal relationship in the video separately. 
Following TimeSformer, we introduce bidirectional Mamba blocks as temporal adapters to replace the vanilla temporal self-attention for improved divided space-time interaction. 
The space attention layers in the TimeSformer remain unchanged for fair comparison.

Here, we use ViM~\cite{zhu2024vim} block as the temporal module, and term the resulting model as TimeMamba.
For consistency, we re-implement the transformer baseline and employ a consistent adaptation method, which involves adding a $\texttt{tanh}$-gating mechanism~\cite{lstm} with an initial value of zero. 
This ensures that the output of the new model matches that of the original model. 
Notably, a standard ViM block has more parameters (slightly more than $6.25 C^2$) than a self-attention block ($4 C^2$), where $C$ is the feature dimension. 
Therefore, we set the expanding ratio $\texttt{E}$ of ViM block to 1, reducing its parameter amount to $3.25 C^2$ for a fair comparison. 
In addition to the vanilla residual connection form used by TimeSformer~\cite{timesformer}, we additionally explored the Frozen-style~\cite{frozenintime} adaption fashion. 
We list blocks with different divided space-time interactions in Figure~\ref{fig:adapter}.
We train the model with 4-frame input using AVION~\cite{zhao2023avion} codebase, with the remaining settings unchanged as \cite{lavila} and \cite{zhao2023avion}. 
The model is initialized with CLIP-B16~\cite{clip} pre-trained via image-text contrastive learning.

\begin{table}[t]

\centering
\scriptsize
\setlength{\tabcolsep}{1mm}{
\begin{tabular}{lcccccccc} 
\toprule
\multirow{2}{*}{ID} & \multirow{2}{*}{Model} & \multirow{2}{*}{Adaptation}  & \multicolumn{3}{c}{mAP} & \multicolumn{3}{c}{nDCG} \\
\cmidrule(lr){4-6}\cmidrule(lr){7-9}
&  & & V2T & T2V & Avg & V2T & T2V & Avg  \\
\midrule
1& EgoVLP~\cite{lin2022egovlp} & Frozen~\cite{frozenintime}, Attn in Time& 19.4 & 13.9 &  16.6 & 24.1 & 22.0 & 23.1   \\
2& EgoVLP~\cite{zhao2023avion} & Frozen~\cite{frozenintime}, Attn in Time& 26.0 & 20.6 &  23.3 & 28.8 & 27.0 & 27.9   \\
3&LaViLa~\cite{lavila}  & Frozen~\cite{frozenintime}, Attn in Time& - & - &  26.0 & - & - & 28.8   \\
\midrule

4& TimeSformer  & Vanilla~\cite{timesformer}, Attn in Time& 29.2 & 21.8 &  25.5 & 30.1  & 27.1 & 28.6   \\
5& TimeSformer  & Frozen~\cite{frozenintime}, Attn in Time & 29.8 & 22.2 &  26.0 & 30.6  & 27.5 & 29.0   \\
\rowcolor{gray!15}
6& TimeMamba (ours) & Vanilla~\cite{timesformer}, Mamba in Time& 30.3 & 22.1 &  26.2 & 30.9  & 27.5 & 29.2   \\
\rowcolor{gray!15}
7& TimeMamba (ours) & Frozen~\cite{frozenintime}, Mamba in Time& \textbf{30.7} & \textbf{22.8} &  \textbf{26.8} & \textbf{31.3}  & \textbf{27.8} & \textbf{29.5}   \\
\rowcolor{gray!15}
8& TimeMamba (ours) & Frozen~\cite{frozenintime}, Mamba in Space-Time& 30.1 & 21.9 &  26.0 & 30.7  & 27.1 & 28.9   \\
\bottomrule

\end{tabular}
}
\vspace{1.5mm}
\caption{Results of zero-shot multi-instance retrieval on EK100~\cite{epic-Kitchens}. 
We compare different models with divided space-time interaction. The additional configuration ``Mamba in space-time'' is used for further comparison. We uniformly sample 4 frames for training and inference.}
\vspace{-2em}
\label{table:zsmir-res}
\end{table}

\begin{table}[t]

\centering
\scriptsize
\setlength{\tabcolsep}{1.75mm}{
\begin{tabular}{lcccccccccc} 
\toprule
 \multirow{4}{*}{Model} & \multicolumn{6}{c}{Multi-instance Retrieval} & \multicolumn{4}{c}{Action Recognition} \\
\cmidrule(lr){2-7} \cmidrule(lr){8-11}
& \multicolumn{3}{c}{mAP} & \multicolumn{3}{c}{nDCG} & Verb & Noun & \multicolumn{2}{c}{Action} \\
\cmidrule(lr){2-4} \cmidrule(lr){5-7} \cmidrule(lr){8-8} \cmidrule(lr){9-9} \cmidrule(lr){10-11}
& V2T & T2V & Avg & V2T & T2V & Avg & Top1 & Top1 & Top1 & Top5  \\
\midrule
EgoVLP~\cite{lin2022egovlp} & 49.9 & \textbf{40.5} & 45.0 & 60.9 & 57.9 & 59.4 & - & - & - & - \\
TimeSformer~\cite{timesformer} & 49.1 & 39.3 & 44.2 & 60.0 & 57.6 & 58.8 & 63.8 & 52.4 & 41.3 & 60.4 \\
\rowcolor{gray!15}
TimeMamba (ours)  & \textbf{50.3} & 40.3 & \textbf{45.3} & \textbf{62.4} & \textbf{59.2} & \textbf{60.9} & \textbf{66.6} & \textbf{53.3} & \textbf{42.8} & \textbf{63.2} \\
\toprule
\end{tabular}
}
\vspace{0.5em}
\caption{Results of fine-tuned multi-instance retrieval and action recognition on EK100~\cite{epic-Kitchens}. We uniformly sample 16 frames for training and inference.}
\vspace{-2em}
\label{table:ft-cls-res}
\end{table}

\begin{figure}[t]
    \centering
    \scriptsize
    \begin{minipage}{0.48 \textwidth}
    \centering
      \begin{tabular}{lcc}
      \toprule
        Model & \#Frame & Accuracy \\
        \midrule
        FrozenBiLM~\cite{yang2022frozenbilm} & 10 & 26.4  \\
        FrozenBiLM~\cite{yang2022frozenbilm} & 90 & 26.9  \\
        InternVideo~\cite{wang2022internvideo} & 15 & 31.6  \\
        InternVideo~\cite{wang2022internvideo} & 90 & 32.0  \\
        \midrule
        TimeSformer~\cite{timesformer} & 16 & 33.6 \\
        TimeSformer~\cite{timesformer} & 128 & 32.9 \\
        TimeSformer~\cite{timesformer} & 8192 & 38.1 \\
        \rowcolor{gray!15}
        TimeMamba (ours) & 16 & 31.9 \\
        \rowcolor{gray!15}
        TimeMamba (ours) & 128 & 33.4 \\
        \rowcolor{gray!15}
        TimeMamba (ours) & 8192 & \textbf{38.7} \\
        \bottomrule
      \end{tabular}
      \captionof{table}{Results of zero-shot long-form video QA on EgoSchema~\cite{mangalam2024egoschema}. The model is trained on Ego4D~\cite{ego4d} with 4 frames and tested on different frames.}
      \label{table:zs-qa-res}

    \end{minipage}\hfill
    \begin{minipage}{0.48 \textwidth}
        \centering
        \vspace{-1em}
        \includegraphics[width=0.8\textwidth]{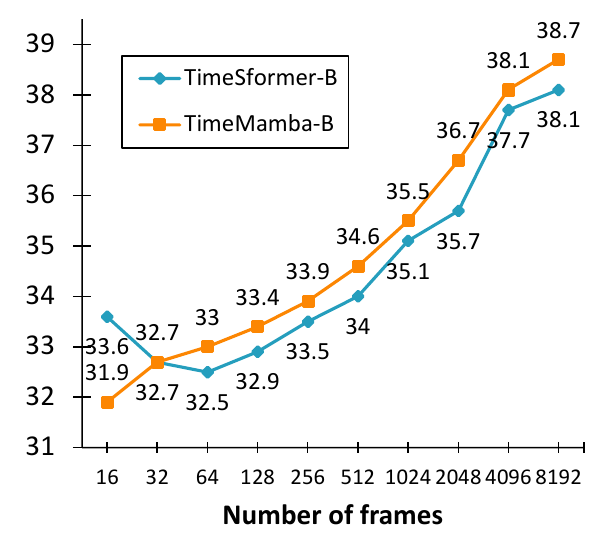}
        \vspace{-1em}
        \caption{The results of using different numbers of testing frames for zero-shot QA on EgoSchema~\cite{mangalam2024egoschema}. The model is trained on Ego4D~\cite{ego4d} with 4 frames.}
        \label{fig:zs-qa-frame}
    \end{minipage}
\end{figure}

\emph{Zero-shot multi-instance retrieval.} 
We first evaluate different models with divided space-time interaction operations in Table~\ref{table:zsmir-res}. 
Our reproduced Frozen style residual connection achieves consistent results to LaviLa~\cite{lavila}.
When comparing vanilla and Frozen~\cite{frozenintime} styles, we observe that the Frozen style consistently yields better results (ID4 vs ID5, ID6 vs ID7). 
Furthermore, under the same adaptation method, the ViM-based temporal module consistently outperforms the attention-based temporal module (ID4 \emph{vs} ID6, ID5 \emph{vs} ID7).  
Notably, the ViM temporal block we used has fewer parameters compared to the temporal self-attention block, highlighting the exceptional parameter utilization and information extraction capabilities of Mamba's selective scanning~\cite{gu2023mamba}. 
Additionally, We go beyond the temporal modeling ability of ViM and validate the space-time ViM block. 
Space-time ViM block replaces the temporal ViM block with joint spatiotemporal modeling over the entire video sequence. 
Surprisingly, we observed that the space-time ViM block, despite introducing global modeling at the spatiotemporal level, actually leads to a decline in performance (ID7 \emph{vs} ID8). We suppose the scanning-based spatial-temporal may damage the spatial feature distribution produced by the pre-trained space attention blocks.


\emph{Fine-tuned multi-instance retrieval and action recognition.} We continue to finetune the pre-trained models with 16 frames on Epic-Kitchen-100~\cite{epic-Kitchens} dataset for multi-instance retrieval and action recognition. In Table~\ref{table:ft-cls-res}, we observe that TimeMamba outperforms TimeSformer by a significant margin. 
In particular, TimeMamba surpasses TimeSformer in the context of verb recognition by 2.8 points, demonstrating its effectiveness in temporal modeling.

\emph{Zero-shot long-form video QA.} 
We conduct further evaluation of the model's long-form video Question Answering performance on EgoSchema~\cite{mangalam2024egoschema}.
As shown in Table~\ref{table:zs-qa-res}, both TimeSformer and TimeMamba, when pre-trained on Ego4D~\cite{ego4d}, outperform the performance of large-scale pre-trained models~\cite{wang2022internvideo,yang2022frozenbilm}.
Additionally, we increase the number of testing frames to explore the effect of ViM block's long-form temporal modeling ability.
As shown in Figure~\ref{fig:zs-qa-frame}, despite both models being pre-trained on 4 frames, the performance of both TimeMamba and TimeSformer steadily improves with increased frames. Meanwhile, significant improvement can be observed when using 8192 frames. 
When the input frame exceeds 32, TimeMamba generally benefits from more frames compared to TimeSformer, indicating the superiority of the temporal ViM block against temporal self-attention.

\begin{figure}[t]
    \centering
    \includegraphics[width=\textwidth]{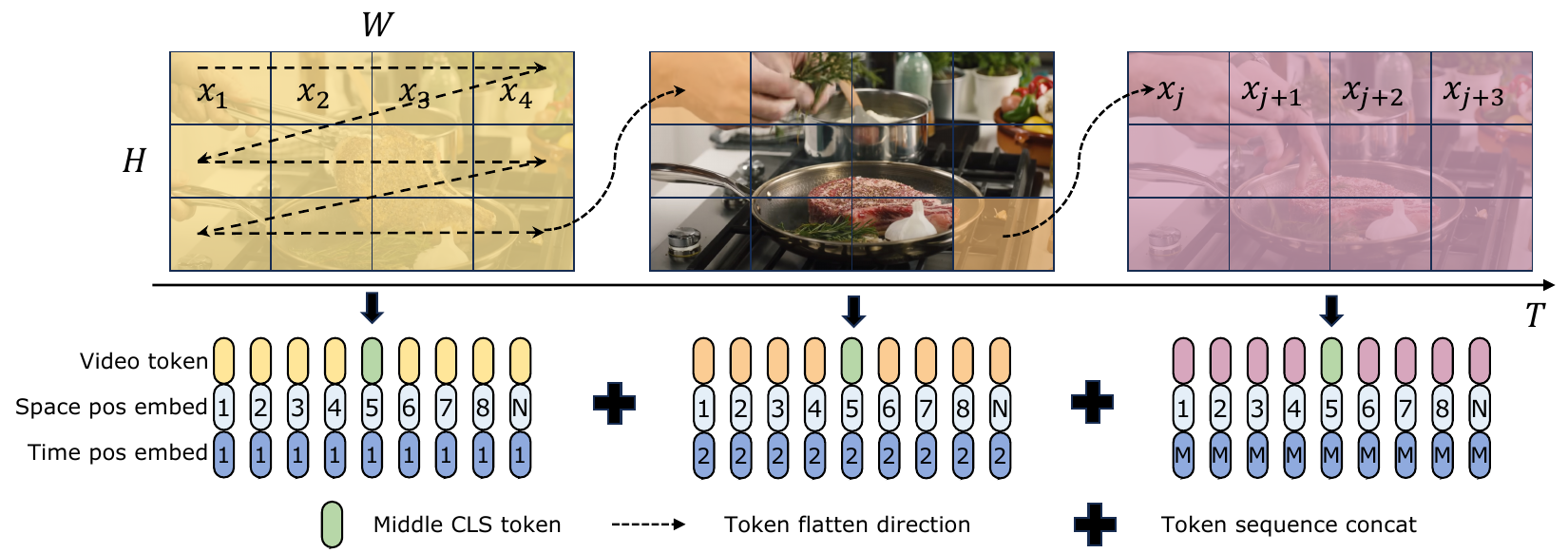}
    \caption{Illustration of the modeling process of our ViViM. We highlight the direction of flattening video tokens through dashed arrows, which determine the process of space-time selective scanning. }
    \label{fig:scan}
\end{figure}

\begin{table}[t]

\centering
\scriptsize
\setlength{\tabcolsep}{1mm}{
\begin{tabular}{lccccccccc} 
\toprule
\multirow{2}{*}{Model} & \multirow{2}{*}{Param} & \multirow{2}{*}{Pretrain}  & \multirow{2}{*}{\#F} & \multicolumn{3}{c}{mAP} & \multicolumn{3}{c}{nDCG} \\
\cmidrule(lr){5-7} \cmidrule(lr){8-10} 
& &  & & V2T & T2V & Avg & V2T & T2V & Avg  \\
\midrule

ViT-B &86M&  CLIP~\cite{clip}, WIT & 4 & 30.95 & 23.15 & 27.05 & 30.95 & 27.65 & 29.30  \\
\midrule
ViT-T &6M& DeiT~\cite{deit}, IN1K~\cite{deng2009imagenet} & 4 & 15.50 & 11.10 &  13.30 & 22.48 & 19.66 & 21.07   \\
ViT-B &86M& DeiT~\cite{deit}, IN1K~\cite{deng2009imagenet} & 4 & 25.08 & 18.49 &  21.79 & 27.80 & 24.87 & 26.34   \\
ViT-T &6M& DeiT~\cite{deit}, IN1K~\cite{deng2009imagenet} & 16 & 20.47 & 15.29 & 17.88 & 25.74 & 22.89 & 24.31   \\
ViT-S &22M& DeiT~\cite{deit}, IN1K~\cite{deng2009imagenet} & 16 & 23.80 & 17.60 & 20.70 & 27.40 & 24.40 & 25.90   \\
\midrule
\rowcolor{gray!15}
ViViM-T  (ours) &7M& DeiT~\cite{deit}, IN1K~\cite{deng2009imagenet} & 16 & 23.31 & 17.21 & 20.26 & 27.40 & 24.30 & 25.80      \\
\rowcolor{gray!15}
ViViM-S (ours) &26M& DeiT~\cite{deit}, IN1K~\cite{deng2009imagenet} & 16 & \textbf{26.00} & \textbf{19.60} & \textbf{22.80} & \textbf{28.20} & \textbf{25.30} & \textbf{26.70}   \\
\bottomrule

\end{tabular}
}
\vspace{2mm}
\caption{Results of zero-shot multi-instance retrieval on EK100~\cite{epic-Kitchens}. We compare ViT with space-time joint attention and our ViViM with space-time joint selective scanning. \#F denotes the frame number for training and inference.}
\vspace{-1em}
\label{table:vivim-mir-res}
\end{table}

\subsection{Mamba for Spatial-Temporal Modeling}

\subsubsection{Tasks and datasets.} 
Finally, we assess the spatial-temporal modeling capability of Mamba. Similar to the previous subsection, we evaluate the model's performance in zero-shot multi-instance retrieval on Epic-Kitchens-100 dataset~\cite{epic-Kitchens}.

\subsubsection{Baseline and competitor.} 
ViViT~\cite{vivit} and TimeSformer~\cite{timesformer} investigated the transformation of ViT with spatial attention into models with spatial-temporal joint attention.
In accordance with these works, we further extend the spatial selective scanning of ViM model~\cite{zhu2024vim} to incorporate space-time selective scanning.
We refer to this extended model as ViViM.
We utilize the ViM model that has been pre-trained on ImageNet-1K~\cite{deng2009imagenet} for initialization. The ViM model incorporates a $\texttt{cls}$ token, which is inserted in the middle of the flattened token sequence.
To convert the ViM model into ViViM, we adopt a straightforward approach illustrated in Figure~\ref{fig:scan}. 
For a given input consisting of $M$ frames, we insert the $\texttt{cls}$ token in the middle of the token sequence corresponding to each frame. Additionally, we add temporal positional embeddings that are initialized to zero for each frame. 
The flattened video sequence is then input into the ViViM model. 
The output of the model is taken by computing the average of $\texttt{cls}$ token from each frame.

\vspace{-1em}
\subsubsection{Results and analysis.}

We further analyze the results of our ViViM on zero-shot multi-instance retrieval.
Table~\ref{table:vivim-mir-res} presents the performance of various spatiotemporal models on zero-shot multi-instance retrieval. 
When comparing ViT and ViViM, both of which are pre-trained on ImageNet-1K~\cite{deng2009imagenet}, we observe that our ViViM outperforms ViT. 
Interestingly, although the performance gap between ViT-S \cite{deit} and ViM-S \cite{zhu2024vim} on ImageNet-1K is slight (79.8 \emph{vs.} 80.5), ViViM-S demonstrates a significant improvement (+2.1 mAP@Avg) over ViT-S on zero-shot multi-instance retrieval. 
This finding suggests that our ViViM is highly effective in modeling long sequences, leading to improved performance.

\section{Efficiency Analysis}
We compare the inference speed of different spatiotemporal models.  This test fixes 196 tokens in spatial dimensions, and continuously increases the number of frames. All tests are performed on a single A100 GPU at half precision. For a fair comparison, all attention blocks are equipped with Flash-attention~\cite{dao2022flashattention,dao2023flashattention2}.

\begin{figure}[t]
    \centering
    \begin{minipage}{0.47 \textwidth}
        \centering
        \includegraphics[width=0.92\textwidth]{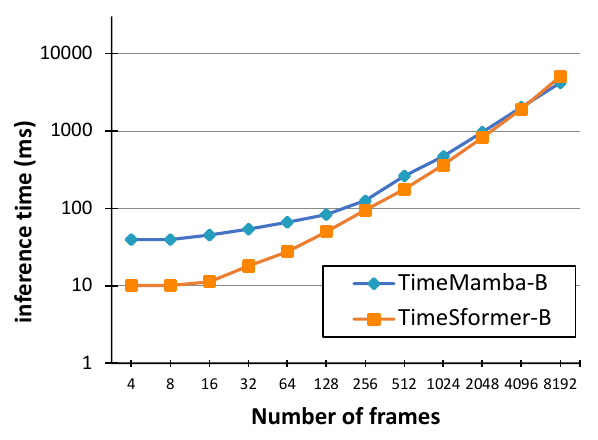}
        \caption{Inference speed of TimeSformer-B and TimeMamba-B with different numbers of frames.}
        \label{fig:eff-timemamba}
    \end{minipage}
    \hspace{1em}
    \begin{minipage}{0.47 \textwidth}
        \centering
        \includegraphics[width=0.92\textwidth]{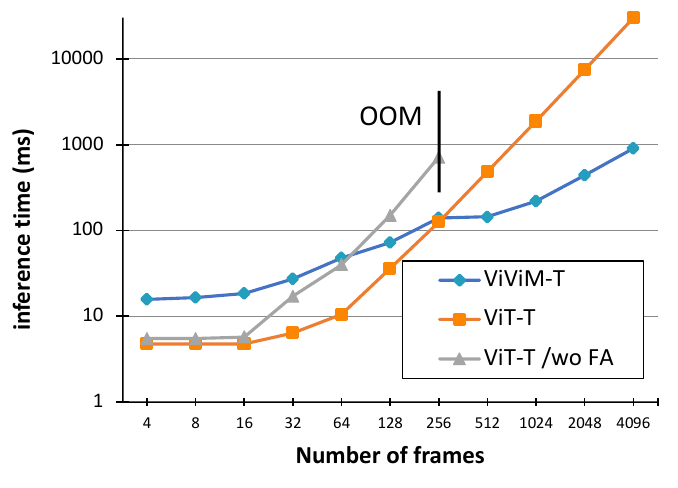}
        \caption{Inference speed of ViT-T and ViViM-T with different numbers of frames.}
        \label{fig:eff-vivim}
    \end{minipage}
\end{figure}

We test the inference speed from 4 frames to 8192 frames and list the test results in Figure~\ref{fig:eff-timemamba} and Figure~\ref{fig:eff-vivim}. Both tables show that Mamba can offer speed advantages over the transformer series models particularly when the number of frames is substantial. In Figure~\ref{fig:eff-vivim}, for fair and comprehensive comparison, we compare ViViM-T with ViT with and without the use of Flash-attention~\cite{dao2022flashattention,dao2023flashattention2}. The comparison of ViViM-T with ViT+Flash-attention is fair because both methods are optimized considering hardware I/O speed. Our ViViM-T becomes more efficient than ViT-T with flash-attention when the input frame number is greater than 256. Without Flash-Attention, ViViM-T is relatively more efficient, surpassing ViT when the frame number is greater than 64. 
For TimeMamba-B in Figure~\ref{fig:eff-timemamba}, when the input is over 8192 frames, the efficiency begins to exceed that of timesformer-B. Since the form of token interaction only differs in time interactions, the efficiency difference is not as significant as the comparison between ViViM and ViT.

\section{Conclusion}
Our comprehensive evaluation of Mamba within the video understanding domain showcases its potential as a viable alternative to traditional transformers. Through the \texttt{Video Mamba Suite}\charimage{example-image}, comprising 14 models/modules across 12 video understanding tasks, we demonstrate Mamba's capability to efficiently handle complex spatial-temporal dynamics, exhibiting both superior performance and promising efficiency-performance trade-offs. These findings not only underline Mamba's suitability for video analysis tasks but also open new avenues for its application in computer vision. Future work could further explore Mamba's adaptability and extend its utility to more complex, multi-modal video understanding challenges.

\clearpage

\renewcommand{\thesection}{A\arabic{section}}
\renewcommand{\thetable}{A\arabic{table}}  
\renewcommand{\thefigure}{A\arabic{figure}}

This supplementary material shows additional details about the models and more experimental results.

\section{More Experimental Results}
In this work, we present a comprehensive study on State Space Models exemplified by Mamba as a versatile alternative in video understanding. We extensively experiment with 4 roles of Mamba in 12 tasks on 13 datasets. Due to the page limitation, in this section, we show experimental results not listed in the main submission.

\subsection{Temporal Action Localization}

Table~\ref{table:tal-thumos-res}, Table~\ref{table:tal-anet-res}, and Table~\ref{table:tal-fineaction-res} present the results of ActionMamba equipped with the ViM~\cite{zhu2024vim} block and our DBM alternative on THUMOS-14~\cite{thumos14}, ActivityNet~\cite{anet} and FineAction~\cite{fineaction} datasets. The superior performance compared to ActionFormer~\cite{actionformer} on all datasets highlights the advantages of bidirectional Mamba in temporal action localization. Furthermore, our DBM block demonstrates superior performance compared to the ViM block. From the results of temporal action localization on all the datasets, we conclude Mamba exhibits stronger capabilities than the Transformer counterpart in the temporal action localization task.
\begin{table}[h]

\centering
\scriptsize
\setlength{\tabcolsep}{0.4mm}{
\begin{tabular}{lccccccc} 
\toprule
Method   & Block  & mAP@0.3 & mAP@0.4 & mAP@0.5 & mAP@0.6 & mAP@0.7  & mAP@Avg \\
\midrule

ActionFormer~\cite{actionformer} & Window Attn~\cite{choromanski2020rethink-attn} & 82.26 & 81.85 & 75.05 & 65.82 & 50.27 & 71.86   \\
\rowcolor{gray!15}
ActionMamba & ViM~\cite{zhu2024vim} & 86.22 & 82.87 & 76.42 & \textbf{66.43} & 50.25 & 72.44   \\
\rowcolor{gray!15}
ActionMamba & DBM (ours) &  \textbf{86.89} & \textbf{83.09} & \textbf{76.90} &   65.91 &   \textbf{50.82} &   \textbf{72.72} \\
\bottomrule
\end{tabular}
}
\vspace{0.5em}
\caption{
Results of temporal action localization on THUMOS-14~\cite{thumos14}. The metric is mean Average Precision (mAP) under multiple tIoU thresholds \{0.3, 0.4, 0.5, 0.6, 0.7\}.
}
\vspace{-5em}
\label{table:tal-thumos-res}
\end{table}
\begin{table}[h]

\centering
\scriptsize
\setlength{\tabcolsep}{2.1mm}{
\begin{tabular}{lccccc} 
\toprule
Method   & Block  & mAP@0.5 & mAP@0.75 & mAP@0.95 & mAP@Avg \\
\midrule

ActionFormer~\cite{actionformer} & Window Attn~\cite{choromanski2020rethink-attn} & 61.47 & 44.61 & 12.73 & 41.19   \\
\rowcolor{gray!15}
ActionMamba & ViM~\cite{zhu2024vim} & 62.31 & 43.17 & 9.65 & 41.77   \\
\rowcolor{gray!15}
ActionMamba & DBM (ours) &  \textbf{62.43} & \textbf{43.49} & \textbf{10.23} &   \textbf{42.02} \\
\bottomrule
\end{tabular}
}
\vspace{0.5em}
\caption{
Results of temporal action localization on ActivityNet~\cite{anet}. The metric is mean Average Precision (mAP) under multiple tIoU thresholds \{0.5, 0.75, 0.95\}.
}
\vspace{-5em}
\label{table:tal-anet-res}
\end{table}

\begin{table}[h]

\centering
\scriptsize
\setlength{\tabcolsep}{2.1mm}{
\begin{tabular}{lccccc} 
\toprule
Method   & Block  & mAP@0.5 & mAP@0.75 & mAP@0.95 & mAP@Avg \\
\midrule

ActionFormer~\cite{actionformer} & Window Attn~\cite{choromanski2020rethink-attn} &43.11 & 27.09 & 5.32 & 27.22   \\
\rowcolor{gray!15}
ActionMamba & ViM~\cite{zhu2024vim} & 44.15 & 28.30 & 6.14 & 28.36   \\
\rowcolor{gray!15}
ActionMamba & DBM (ours) &  \textbf{45.44} & \textbf{28.82} & \textbf{6.79} &   \textbf{29.04} \\
\bottomrule
\end{tabular}
}
\vspace{0.5em}
\caption{
Results of temporal action localization on FineAction~\cite{fineaction}. The metric is mean Average Precision (mAP) under multiple tIoU thresholds \{0.5, 0.75, 0.95\}.
}
\vspace{-1em}
\label{table:tal-fineaction-res}
\end{table}

\subsection{Temporal Action Segmentation}
Table~\ref{table:tas-breakfast-res} and Table~\ref{table:tas-50salads-res} show the performances of the temporal action segmentation task on the Breakfast and 50Salads datasets. On the Breakfast dataset (Table~\ref{table:tas-breakfast-res}), Mamba-based methods achieve significantly stronger performances. 
On the 50Salads dataset (Table~\ref{table:tas-50salads-res}), Transformer-based method with encoder-decoder structure, ASFormer~\cite{asformer}, gets better results in terms of Accuracy, Edit score, and F1@10, while Mamba-based methods, ASMamba, perform better on F1@25. 
We also observe that ASMamba outperforms encoder-only ASFormer, highlighting the strong temporal encoding capability of the Mamba-based model. These results further reveal that the encoder-decoder structure may be a better choice compared with the encode-only structure in some scenarios.


\begin{table}[t]
\centering
\scriptsize
\setlength{\tabcolsep}{2.4mm}{
\begin{tabular}{lcccccc} 
\toprule
Method   & Block  & Acc & Edit & F1@10 & F1@25 & F1@50 \\
\midrule
MS-TCN~\cite{ms-tcn} & Dilated Conv, Enc & 66.3 & 61.7 & 52.6 & 48.1 & 37.9   \\
\midrule
ASFormer~\cite{asformer} & Window Attn~\cite{choromanski2020rethink-attn}, Enc-Dec & 73.5 & 75.0 & 76.0 & 70.6 & 57.4   \\
\rowcolor{gray!15}
ASMamba & ViM~\cite{zhu2024vim}, Enc & \textbf{75.8} & \textbf{76.9} & \textbf{78.7} & \textbf{73.8} & \textbf{61.6}   \\
\rowcolor{gray!15}
ASMamba & DBM (ours), Enc & 73.5 & 75.8 & 77.4 & 72.0 & 59.3   \\
\bottomrule
\end{tabular}
}
\vspace{0.5em}
\caption{
Results of temporal action segmentation on Breakfast~\cite{breakfast} dataset. 
The metrics are accuracy, edit distance~\cite{edit-distance}, and instance-wise F1 under multiple tIoU thresholds \{0.1, 0.25, 0.5\}. $\dagger$ denotes our reproduced results with its official code.}
\vspace{-1em}
\label{table:tas-breakfast-res}
\end{table}

\begin{table}[t]
\centering
\scriptsize
\setlength{\tabcolsep}{2.4mm}{
\begin{tabular}{lcccccc} 
\toprule
Method   & Block  & Acc & Edit & F1@10 & F1@25 & F1@50 \\
\midrule
MS-TCN~\cite{ms-tcn} & Dilated Conv, Enc & 80.7 & 67.9 & 76.3 & 74.0 & 64.5   \\
\midrule
ASFormer~\cite{asformer} & Window Attn~\cite{choromanski2020rethink-attn}, Enc-Enc & 77.0 & 64.4 & 73.0 & 70.5 & 60.3   \\
ASFormer~\cite{asformer} & Window Attn~\cite{choromanski2020rethink-attn}, Enc-Dec & \textbf{85.6} & \textbf{79.6} & \textbf{85.1} & 73.4 & \textbf{76.0}   \\
\rowcolor{gray!15}
ASMamba & ViM~\cite{zhu2024vim}, Enc & 85.4 & 77.7 & 84.5 & \textbf{83.4} & 74.7   \\
\rowcolor{gray!15}
ASMamba & DBM (ours), Enc & 83.7 & 75.2 & 82.5 & 80.5 & 72.5   \\
\bottomrule
\end{tabular}
}
\vspace{0.5em}
\caption{
Results of temporal action segmentation on 50salads~\cite{50salads} dataset. 
The metrics are accuracy, edit distance~\cite{edit-distance}, and instance-wise F1 under multiple tIoU thresholds \{0.1, 0.25, 0.5\}. $\dagger$ denotes our reproduced results with its official code.}
\vspace{-1em}
\label{table:tas-50salads-res}
\end{table}

\subsection{Dense Video Captioning}
We also present the localization metrics (Recall and Precision) of dense video captioning in Table~\ref{table:dvc-anet-res} and Table~\ref{table:dvc-youcook-res}. 
The results demonstrate that the Mamba-based temporal encoder produces improved position representations and enhances the localization capability, which aligns with its superior performance on the temporal action localization task.

\begin{table}[h]
\centering
\scriptsize
\setlength{\tabcolsep}{1.2mm}{
\begin{tabular}{lcccccccc} 
\toprule
  Method & Block  &  B-4 & METEOR & ROUGE-L & CIDER & SODA  & Recall & Precision  \\
\midrule
PDVC~\cite{wang2021pdvc} & DeformAttn~\cite{deformable-detr} & 1.75 & 6.73 & 14.65 & 26.07 & \textbf{5.47} & 51.7 & 56.1  \\
\rowcolor{gray!15}
PDVC~\cite{wang2021pdvc} & ViM~\cite{zhu2024vim} & 1.68 & 6.92 & 14.72 & 26.26 & 5.33 & \textbf{53.1} & \textbf{56.3} \\
\rowcolor{gray!15}
PDVC & DBM (ours) & \textbf{1.76} & \textbf{7.16} &\textbf{14.83} &\textbf{26.77} & 5.27 & 52.4 & \textbf{56.3}\\
\bottomrule
\end{tabular}
}
\vspace{0.5em}
\caption{
Results of dense video captioning on ActivityNet~\cite{anet}. 
The metrics include BLEU-4~\cite{papineni2002bleu}, METEOR~\cite{banerjee2005meteor}, CIDEr~\cite{vedantam2015cider} SODA\_c~\cite{fujita2020soda}, Recall and Precision.
}
\label{table:dvc-anet-res}
\end{table}

\begin{table}[h]
\centering
\scriptsize
\setlength{\tabcolsep}{1.2mm}{
\begin{tabular}{lcccccccc} 
\toprule
  Method & Block  &  B-4 & METEOR & ROUGE-L & CIDER & SODA  & Recall & Precision  \\
\midrule
PDVC~\cite{wang2021pdvc} & DeformAttn~\cite{deformable-detr} & 0.73 & 4.25 & 9.31 & 20.48 & 4.02 & 23.0 & 31.1  \\
\rowcolor{gray!15}
PDVC~\cite{wang2021pdvc} & ViM~\cite{zhu2024vim} & 0.71 & 4.32 & 9.57 & 20.59 & 4.09 & 24.1 & \textbf{33.0} \\
\rowcolor{gray!15}
PDVC & DBM (ours) & \textbf{0.86} & \textbf{4.44} & \textbf{9.64} &\textbf{22.11} & \textbf{4.32} & \textbf{25.2} & 32.4\\
\bottomrule
\end{tabular}
}
\vspace{0.5em}
\caption{
Results of dense video captioning on YouCook2~\cite{zhou2018youcook2}. 
The metrics include BLEU-4~\cite{papineni2002bleu}, METEOR~\cite{banerjee2005meteor}, CIDEr~\cite{vedantam2015cider}, SODA\_c~\cite{fujita2020soda}, Recall and Precision.
}
\vspace{-1em}
\label{table:dvc-youcook-res}
\end{table}

\subsection{Action Recognition}

Furthermore, we present the results of action recognition on Kinetics-400~\cite{kinetics}, as demonstrated in Table~\ref{table:ar-k400-res}. 
ViViM-T and ViViM-S achieve 77.4\% and 80.1\% Top-1 accuracy, respectively. Compared with Transformer-based models with tiny size (\emph{e.g.} 28M VideoSwin-T~\cite{video-swin}), ViViM-T achieves a competitive performance with significantly fewer parameters (7M). 
One can still observe a performance gap between ViViM and models with dedicated spatial/temporal/spatio-temporal module designs (\emph{e.g.} UniFormer~\cite{uniformer}), and we leave the design of SSM-based modules for video understanding as future work.

\begin{table}[h]

\centering
\scriptsize
\setlength{\tabcolsep}{1.2mm}{
\begin{tabular}{lccccccc} 
\toprule
Method   & Arch  & Pretrain & Input Size & Param & GFLOPs & Top-1 & Top-5   \\
\midrule
SlowFast$_{\text{R101+NL}}$~\cite{slowfast} & CNN & - & 80$\times$ $224^2$& 60 & 234$\times$3$\times$10 &   79.8 &  93.9  \\
X3D-XL~\cite{feichtenhofer2020x3d} & CNN & - & 16$\times$ $312^2$& 20 & 194$\times$3$\times$10 &   80.4 &  94.6  \\
VideoSwin-T~\cite{video-swin} & Trans & IN1K~\cite{deng2009imagenet} & 32$\times$ $224^2$& 28 & 88$\times$3$\times$4 &   78.8 &  93.6  \\
VideoSwin-B~\cite{video-swin} & Trans & IN1K~\cite{deng2009imagenet} & 32$\times$ $224^2$& 88 & 88$\times$3$\times$4 &   80.6 &  94.5 \\
MViTv1-B~\cite{mvit} & Trans+CNN & - & 32$\times$ $224^2$& 37 & 70$\times$1$\times$5 &   80.2 &  94.4  \\
UniFormer-S~\cite{uniformer} & Trans+CNN & IN1K~\cite{deng2009imagenet} & 16$\times$ $224^2$& 21 & 42$\times$1$\times$4 &   80.8 &  94.7  \\
STAM~\cite{sharir2021stam} & Trans & IN22K~\cite{deng2009imagenet} & 64$\times$ $224^2$& 121 & 1040$\times$1$\times$1 &   79.2 &  -  \\
TimeSformer-L~\cite{timesformer} & Trans & IN22K~\cite{deng2009imagenet} & 96$\times$ $224^2$& 311 & 2380$\times$3$\times$1 &   80.7 &  94.7  \\
ViViT-L~\cite{vivit} & Trans & IN22K~\cite{deng2009imagenet} & 16$\times$ $224^2$& 311 & 3992$\times$3$\times$4 &   81.3 &  94.7  \\
\midrule
\rowcolor{gray!15}
ViViM-T (ours) & SSM & IN1K~\cite{deng2009imagenet} & 16$\times$ $224^2$& 7 & 17$\times$3$\times$4 &   77.4 &  92.8  \\
\rowcolor{gray!15}
ViViM-S (ours) & SSM & IN1K~\cite{deng2009imagenet} & 16$\times$ $224^2$& 26 & 68$\times$3$\times$4 &   80.1 &  94.1  \\
\bottomrule
\end{tabular}
}
\vspace{0.5em}
\caption{
Comparison with the previous methods on Kinetics-400~\cite{kinetics}.
}
\vspace{-1em}
\label{table:ar-k400-res}
\end{table}

\section{ViM and DBM}

In this section, we provide a detailed description of our explored DBM block and its differences from the ViM block. 

\textbf{ViM block.} Given an input sequence $\mathbf{x} \in \mathbb{R}^{N\times d}$, where $N$ represents the sequence length and $d$ represents the feature dimension, the ViM block initially expands the dimension to $d \cdot \texttt{E}$, resulting in two hidden state sequences, namely $\mathbf{x}_s$ and $\mathbf{x}_g$. The first sequence, $\mathbf{x}_s$, is intended for scanning, while the second sequence, $\mathbf{x}_g$, is used for gating. The selective scanning layer scans $\mathbf{x}_s$ in both directions using two distinct parameters. Finally, $\mathbf{x}_g$ is employed to gate the bidirectionally scanned features, and the output feature is obtained through the average of the two gated features.

\textbf{DBM block.} In the DBM block, given an input sequence $\mathbf{x} \in \mathbb{R}^{N\times d}$, where $N$ represents the sequence length and $d$ represents the feature dimension, the process is similar to the ViM block. Initially, the DBM block expands the dimension to $d \cdot \texttt{E}$, resulting in two hidden state sequences, namely $\mathbf{x}_s$ and $\mathbf{x}_g$.
However, in the DBM block, the forward and backward features are separated along the channel dimension, resulting in four hidden state sequences: $\mathbf{x}_s^{forward}, \mathbf{x}_g^{forward}, \mathbf{x}_s^{backward}, \mathbf{x}_g^{backward} \in \mathbb{R}^{N \times d\cdot \frac{\texttt{E}}{2}}$.
The selective scanning layer scans $\mathbf{x}_s^{forward}$ and $\mathbf{x}_g^{forward}$ using shared parameters. Finally, $\mathbf{x}_g^{forward}$ and $\mathbf{x}_g^{backward}$ are used to gate the both bidirectionally scanned features, respectively. The output feature is obtained by concatenating and projecting the two gated features.

\textbf{Analysis.} Considering $\texttt{E}=2$, the ViM block and DBM block exhibit differences in terms of the number of parameters utilized and the scanning context for static and dynamic modeling. With Mamba~\cite{gu2023mamba} serving as the baseline, Table~\ref{table:block-analysis} presents the disparity in parameters, context, and time cost between ViM~\cite{zhu2024vim} and DBM. 
Compared to the ViM block, the DBM block provides static direction separation and reduces the capacity for dynamic modeling, making it more compatible with certain downstream datasets and tasks, such as temporal action localization. 
Additionally, halving the scanning context results in a corresponding halving of the (training and inference) time cost, consistent with the vanilla Mamba~\cite{gu2023mamba} block.

\begin{table}[t!]
    \centering
    \scriptsize
    \setlength{\tabcolsep}{2.5mm}{
    \begin{tabular}{lccccc}
    \toprule
       Block & Param (sta) & Param (dyn) & Memory (sta) &  Memory (dyn) & Time cost \\
       \midrule
       Mamba~\cite{gu2023mamba} & 100\% & 100\% & 100\% & 100\% & 100\%\\ 
       \midrule
       ViM~\cite{zhu2024vim}  & 100\% & 200\% & 100\% & 200\% & 200\% \\
        \rowcolor{gray!15}
       DBM (ours)  & 100\% & 100\% & 100\% & 100\% & 100\% \\
       \bottomrule
    \end{tabular}
    }
    \vspace{0.5em}
    \caption{Parameters, context and cost time comparison between Mamba~\cite{gu2023mamba}, ViM~\cite{zhu2024vim} and DBM. ``sta'' denotes the static projection and ``dyn'' is dynamic scanning.}
    \label{table:block-analysis}
\end{table}

\section{Hyperparameter Sensitivity}

Additionally, we conducted an analysis of the hyperparameter sensitivity of the Mamba series models. In most experiments, the training hyperparameters were found to be insensitive. We simply replaced the transformer block with the Mamba-based block for most tasks. However, for video temporal grounding, we observed that a larger learning rate yielded better optimization results. Furthermore, increasing the loss weight for video-text alignment loss facilitated the model's convergence. We postulate that these adjustments are related to the distinction between the scanning mechanism and global interaction, especially for mutli-modal aggregation.

%
%
\bibliographystyle{splncs04}
\bibliography{egbib}
\end{document}